\documentclass[runningheads]{llncs}

\usepackage{eccv}

\usepackage{eccvabbrv}

\usepackage{graphicx}
\usepackage{booktabs}
\usepackage{makecell}
\usepackage{multirow}

\usepackage[accsupp]{axessibility}  

\usepackage{xspace}  
\newcommand{\method}{DriftScope\xspace}

\usepackage{hyperref}

\usepackage{orcidlink}

\begin{document}

\title{\method : Measuring The Hidden Effects of Diffusion Model Adaptation} 

\titlerunning{\method}

\author{Héctor Laria$^*$\inst{1,2}\orcidlink{0009-0008-6253-4709}\and
Yiping Han$^*$\inst{1,2}\orcidlink{0009-0007-2589-576X}\and
Julian D. Santamaria\inst{1,2}\orcidlink{0009-0007-7287-5761}\and \\
Kai Wang\inst{3,4}$^\dag$\orcidlink{0000-0002-9605-8279} \and
Bogdan Raducanu\inst{1,2}\orcidlink{0000-0003-3648-8020}\and \\
Joost van de Weijer\inst{1,2}\orcidlink{0000-0002-9656-9706}\and 
Alexandra Gomez-Villa\inst{1,2}\orcidlink{0000-0003-0469-3425}}

\authorrunning{H.~Laria et al.}

\institute{Computer Vision Center, Barcelona, Spain \and
Universitat Autonoma de Barcelona, Barcelona, Spain \and
Program of Computer Science, City University of Hong Kong (Dongguan), China \and
City University of Hong Kong, HK SAR, China \\
\email{\{{hlaria, yhan, jsantamaria, bogdan, joost, agomezvi\}@cvc.uab.es}, {kai.wang}@cityu-dg.edu.cn}}
\maketitle
\setcounter{footnote}{0}
\renewcommand{\thefootnote}{\fnsymbol{footnote}}
\footnotetext{$^*$ Equal Contribution.}
\footnotetext{$^\dag$ Corresponding Author.}
\footnotetext{Project page: \url{https://hyping111.github.io/DriftScope/}}

\begin{abstract}

Adapting pre-trained text-to-image diffusion models, whether to learn new visual concepts or erase unwanted ones, is routinely evaluated on its intended effects alone. We argue this framing is incomplete. Through sparse autoencoder analysis and zero-shot classification, we demonstrate that adaptation systematically damages semantically unrelated concepts in ways that aggregate metrics structurally cannot surface: when damage is severe enough for FID and KID to respond, the model is already nearly unusable; when the model remains functional, FID and KID stay flat while specific classes silently suffer worst-case zero-shot accuracy drops of up to 18.9 points and concept-level distributions shift dramatically. This pattern appears at both ends of the adaptation spectrum (concept customization and concept unlearning), suggesting it is a systematic consequence of weight-level modification rather than an artifact of any particular method. To surface this hidden drift before deployment, we introduce \method, a prompt-level diagnostic tool that takes any two model checkpoints and returns a ranked list of tokens whose visual concepts have shifted most between them. \method optimizes a soft prompt to attribute drift at the token level without requiring access to real data or model internals. The result is an interpretable, concept-level audit that aggregate evaluation cannot provide.

  \keywords{Diffusion models  \and Model evaluation \and Model diffing}
\end{abstract}

\section{Introduction}
\label{sec:intro}

Adapting foundation models to specific needs has become a central challenge in modern AI deployment, and weight-level adaptation is among the most widely adopted solutions. Whether customizing text-to-image models to new visual concepts~\cite{ruiz2023dreambooth} or removing unsafe content through unlearning~\cite{gandikota2023erasing}, practitioners routinely modify pre-trained weights to serve specific needs. These operations are treated as benign optimizations, evaluated mostly on their intended effects: Does the customized model generate the new concept? Does the unlearned model avoid harmful content?

We argue this framing is incomplete. While adaptation achieves its proximal goals, it simultaneously degrades the model in ways current evaluation misses~\cite{laria2024openworldforgetting}. Standard metrics (such as FID, KID, prompt-specific evaluations) focus on aggregate quality and target-task performance, missing fine-grained distributional shifts. As a result, adapted models may lose entire categories of concepts and exhibit dramatic perceptual shifts on unrelated tasks, all while passing standard benchmarks. These ``hidden costs'' arise across opposite ends of the adaptation spectrum: concept customization, which \emph{adds} a new concept, and concept unlearning, which \emph{removes} one, suggesting this is a systematic consequence of weight-level adaptation, not an artifact of any particular method.

The core problem is granularity. Standard metrics measure aggregate quality; they are structurally blind to concentrated damage on specific concepts. What practitioners need is not another aggregate score, but a tool that answers a precise question: \textit{Given this prompt, which words have their generated concepts most affected by adaptation?}



In this paper, we first show that this drift is real and systematic through two complementary analyses: sparse autoencoders (SAEs) to surface concept-level shifts in the generative distribution, and zero-shot classification via GDC~\cite{qi2024simple} to confirm these shifts are reflected in the model's latent space. Together they reveal that adaptation inflicts damage on the model's capabilities while leaving aggregate metrics intact, exactly the failure mode that FID-based evaluation cannot surface. Building on this evidence, we introduce \method, a diagnostic tool that takes any prompt and two model checkpoints (a base model and an adapted variant) and returns a word-level drift report identifying where semantic coverage has eroded. \method generates a prompt using the base model and feeds it through both checkpoints simultaneously, maximizing the distance between the cross-attention maps produced by each model. Words whose cross-attention activations diverge most between the two models are identified as the concepts most affected by adaptation, often surfacing non-obvious collateral damage beyond the adapted target.

In summary, this  paper makes the following contributions:
\begin{itemize}



    \item We show that concept customization and concept unlearning, both induce collateral damage on semantically unrelated concepts. Aggregate metrics such as FID and KID either remain flat while damage accumulates silently, or only respond once the model has become nearly unusable, revealing a structural blind spot in standard evaluation.
    \item We establish this through sparse autoencoder analysis, which surfaces concept-level shifts in the generative distribution, and zero-shot classification, which confirms these shifts in the model's latent space.
    \item We introduce \method, a prompt-level diagnostic tool that returns a ranked blacklist of tokens whose visual concepts have drifted most between any two checkpoints, via differentiable cross-attention divergence maximization.

\end{itemize}

\section{Related Work}
\label{sec:related}

\subsubsection{Divergent representations in T2I models.} 
Divergent representations refer to systematic discrepancies between what a generative model is expected to produce and what it actually generates, whether measured against a reference data distribution or against another model checkpoint. Bohacek \etal~\cite{bohacek2025blindspots} formalize this as \textit{conceptual blindspots}, using sparse autoencoders over DINOv2 features to identify concepts that are suppressed or exaggerated in generated images relative to training data. In a complementary direction, Tong \etal~\cite{tong2023mass} mine for \textit{erroneous agreements} in CLIP and use a language model to surface systematic failure patterns that propagate through downstream T2I systems. Dunlap \etal~\cite{dunlap2025discovering} shift the lens to cross-model comparison, introducing COMPCON, an evolutionary search algorithm that discovers visual attributes differing between two \emph{independently trained} T2I models. While effective for model selection, COMPCON relies on iterative calls to external VLMs and LLMs and is non-differentiable by design; because it compares models that differ in architecture, training data, and procedure simultaneously, the divergences it surfaces cannot be attributed to any specific cause. \method targets a more precise setting (both checkpoints share the same architecture and weights, differing only in a targeted adaptation) and is entirely self-contained, using gradient backpropagation through two frozen checkpoints to produce token-level drift scores directly actionable for pre-deployment auditing.

\subsubsection{Adversarial Prompting in Text-to-Image Models.}
A line of work has studied how carefully crafted prompts can expose weaknesses in T2I diffusion models. Liu \etal~\cite{liu2024discovering} propose SAGE, an adversarial search method that systematically explores the prompt and latent space to surface failure modes. Tsai \etal~\cite{tsai2024ringabell} introduce Ring-A-Bell, a model-agnostic red-teaming scheme that constructs adversarial prompts to bypass concept removal mechanisms. Both approaches optimize prompts to probe a \emph{single} model's behavior, either to find failures or circumvent erasure. DEXTER~\cite{carnemolla2025dexter} uses differentiable prompt optimization (a masked LM with Gumbel-Softmax relaxation) to maximize neuron activations in a visual classifier, relying on external VLMs for global explanations. \method shares the same prompt construction infrastructure but uses it as a self-contained \emph{measurement instrument} to compare two generative checkpoints, localizing tokens whose visual concepts have shifted as a result of adaptation.

\subsubsection{Continual Learning and Catastrophic Forgetting in Generative Models.}
Catastrophic forgetting~\cite{masana2022class} has been studied in text-to-image models primarily through the lens of continual customization. Methods such as C-LoRA~\cite{smith2024continualdiffusion} and FL2T~\cite{kaushik2026forget} address the sequential setting where concepts are added one after another; Zaj\c{a}c~\etal~\cite{Zajac2023ExploringCL} further benchmark continual learning baselines within diffusion models, identifying timestep-dependent forgetting patterns. These works share the assumption that forgetting is a \emph{known problem} to be mitigated across a sequence of adaptation steps. Our work addresses a more fundamental prerequisite: in a \emph{single} adaptation step, \method does not prevent forgetting. It surfaces it, providing a concept-level audit that is a necessary precondition for any mitigation strategy.
This builds directly on prior work~\cite{laria2024openworldforgetting}, which first characterized open-world forgetting in diffusion model customization via zero-shot classification and appearance drift analysis; here we extend that investigation to concept unlearning and introduce a diagnostic tool to identify \emph{which} concepts are affected.

\subsubsection{Model diffing.}
Model diffing refers to systematically comparing two model checkpoints to localize what has changed between them. Lindsey \etal~\cite{lindsey2025crosscoder} introduce crosscoders, sparse autoencoders trained simultaneously on residual stream activations of two language model checkpoints, producing features that cluster into shared, base-model-specific, and fine-tuned-model-specific groups. Bricken \etal~\cite{bricken2024modeldifffinetuning} tracks similar changes by fine-tuning an existing SAE dictionary on the post-adaptation model, measuring how individual features rotate or are suppressed. Both methods operate over internal activations of language models. \method shares the diagnostic intent but targets a different modality: rather than probing residual stream features, it compares cross-attention maps in text-to-image diffusion models and surfaces concept-level drift at the token level, requiring no access to model internals beyond standard attention outputs.

\section{Method}
\label{sec:method}

\subsection{Preliminaries}

\subsubsection{Diffusion models.} 
Diffusion models~\cite{sohl2015deep,ho2020denoising,song2020score} are a class of generative models that learn to synthesize data by reversing a gradual noising process. The forward process progressively adds Gaussian noise to data $\mathbf{x}_0 \sim q(\mathbf{x}_0)$ over $T$ timesteps according to a variance schedule $\{\beta_t\}_{t=1}^T$, producing increasingly noisy samples $\mathbf{x}_t$ until $\mathbf{x}_T \sim \mathcal{N}(\mathbf{0}, \mathbf{I})$. The reverse process learns to denoise these samples by training a neural network $\epsilon_\theta(\mathbf{x}_t, t)$ to predict the noise added at each timestep, optimizing the variational lower bound: $\mathbb{E}_{t, \mathbf{x}_0, \epsilon}\left[\|\epsilon - \epsilon_\theta(\mathbf{x}_t, t)\|^2\right]$. At inference time, samples are generated by iteratively denoising from pure noise $\mathbf{x}_T$ through the learned reverse process.

\subsubsection{Diffusion model adaptation.}
We study two paradigms through which practitioners routinely modify pre-trained text-to-image diffusion models.
\begin{itemize}
    \item \textbf{Concept unlearning} aims to surgically remove a target concept from a model's generative distribution without retraining from scratch. We evaluate four methods: Ablating Concepts (AC)~\cite{gandikota2023erasing}, which fine-tunes the model to redirect the target concept toward an anchor; EraseDiff~\cite{wu2025erasing}, which formulates unlearning as a constrained optimization over the denoising objective; MACE~\cite{lu2024mace}, which uses closed-form weight updates via a multi-concept erasure framework; and SPM~\cite{lyu2023onedimensional}, which trains lightweight concept-specific modules to suppress target activations.
    \item \textbf{Model customization} adapts a pre-trained model to faithfully generate a specific user-provided visual concept from only a handful of reference images. We evaluate four methods: DreamBooth~\cite{ruiz2023dreambooth}, which fine-tunes the full model with a class-preservation loss; Custom Diffusion~\cite{kumari2022customdiffusion}, which restricts fine-tuning to cross-attention key and value projections; BOFT~\cite{liu2024boft}, which reparameterizes weight updates via butterfly orthogonal transformations; and SVDiff~\cite{han2023svdiff}, which fine-tunes singular values of weight matrices. We evaluate across three base architectures (SD 1.5, SD 2.1, and SD 3.5) to assess whether the observed drift patterns are model-dependent.
\end{itemize}

Throughout this paper, we refer to the \textbf{base model} as the original pre-trained text-to-image diffusion model (\eg, Stable Diffusion 3.5), and the \textbf{modified model} as the same architecture after adaptation. 
In all cases, we compare generations from the base and modified models using identical prompts and random seeds to isolate the effects of the adaptation.

\subsection{Model Adaptation Induces Hidden Distributional Drift}

\subsubsection{Concept-Level Drift}

To probe whether model adaptation alters the generative distribution beyond its intended target, we analyze concept-level shifts using the pipeline introduced by Bohacek \etal~\cite{bohacek2025blindspots}. Their method trains a sparse autoencoder (SAE) over DINOv2 image features to decompose image embeddings into approximately interpretable concept activations, enabling quantitative comparison of concept prevalence across image sets. Originally designed to compare model-generated images against real training data, we repurpose this diagnostic for a different setting: comparing two sets of images generated by the \emph{same architecture} under two different weight configurations. This eliminates the need for real reference images and allows drift to be attributed directly to the adaptation.

We sample 10{,}000 prompts from DiffusionDB and generate paired images from the base model $\mathcal{M}$ and its adapted counterpart $\mathcal{M}'$ using identical prompts and noise seeds. Let $\varepsilon_k(x)$ denote the SAE activation corresponding to concept $k$. For each concept, we compute a drift score:
\begin{equation}
\omega(k) 
=
\sigma\!\left(
\mathbb{E}_{x' \sim \mathcal{M}'}[\varepsilon_k(x')]
-
\mathbb{E}_{x \sim \mathcal{M}}[\varepsilon_k(x)]
\right),
\end{equation}
where $\sigma(\cdot)$ is the sigmoid function and $\omega(k)=0.5$ indicates perfect alignment between the two distributions.

\begin{figure}[tb]
    \centering
    \includegraphics[width=0.24\textwidth]{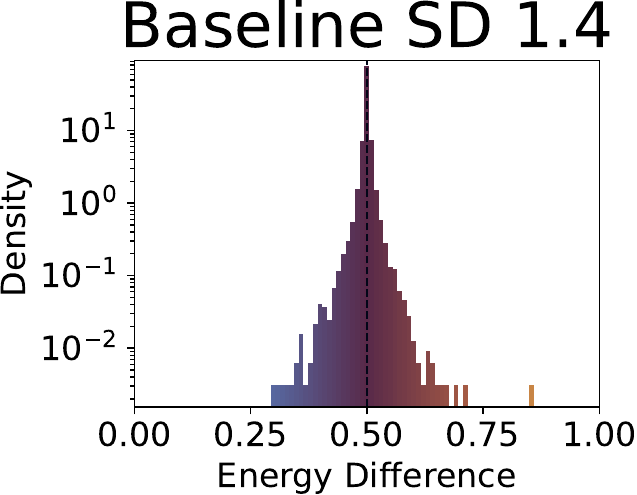}\hfill
    \includegraphics[width=0.24\textwidth]{figures//blindspots_hist/ac_nudity}\hfill
    \includegraphics[width=0.24\textwidth]{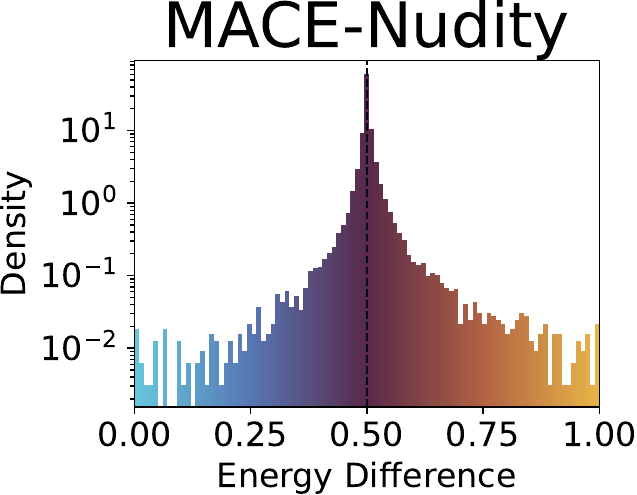}\hfill
    \includegraphics[width=0.24\textwidth]{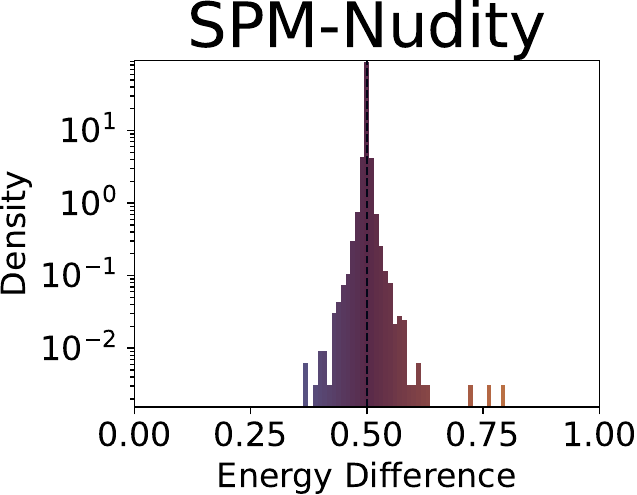}

    \vspace{0.5em}

    \includegraphics[width=0.24\textwidth]{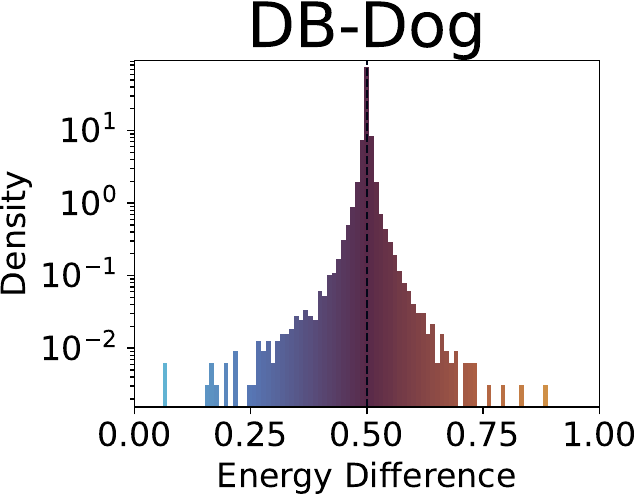}\hfill
    \includegraphics[width=0.24\textwidth]{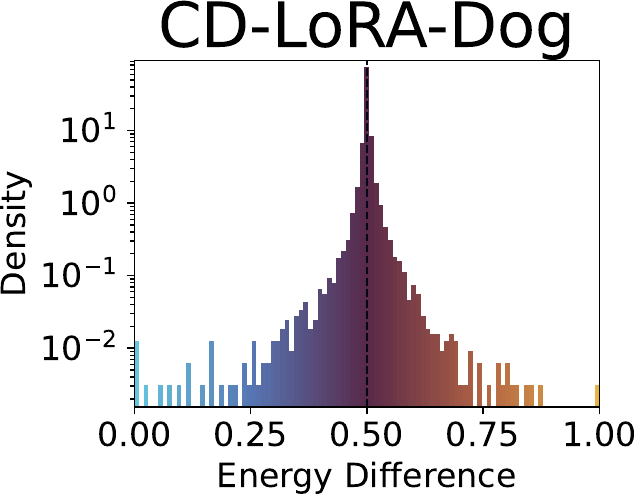}\hfill
    \includegraphics[width=0.24\textwidth]{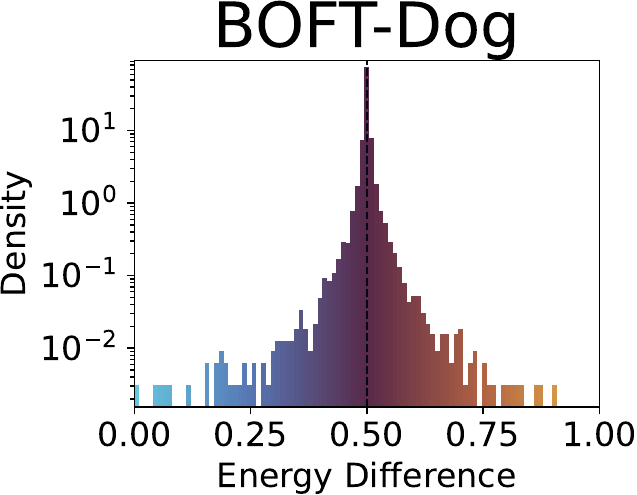}\hfill
    \includegraphics[width=0.24\textwidth]{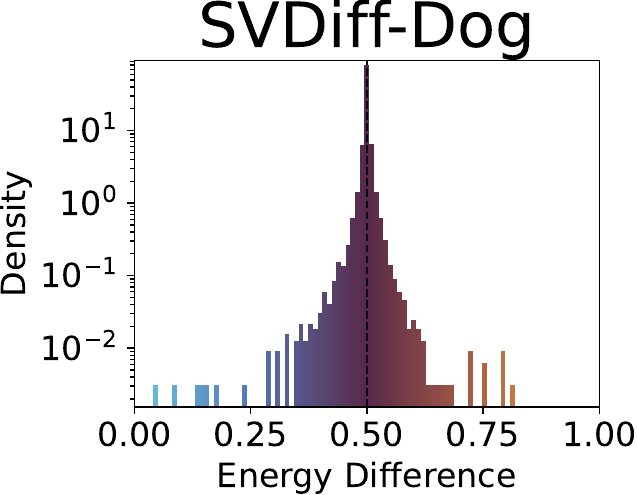}



    \caption{Distribution of SAE drift scores $\omega(k)$ across concepts for 
multiple adaptation paradigms. \textbf{First panel (top-left):} baseline 
control (same architecture, same prompts, different random seeds) 
showing that stochastic sampling variance alone concentrates sharply around $0.5$. All remaining panels show adapted variants, where heavy-tailed behavior indicates systematic concept-level drift beyond the adapted target. Top row: concept unlearning (nudity target). Bottom row: concept customization (dog concept). 
See Appendix~\ref{app:blindspots_full} for the full set of distributions across all evaluated methods.
}
    \label{fig:grid_blindspots}
    \vspace{-4mm}
\end{figure}

Figure~\ref{fig:grid_blindspots} shows the distribution of $\omega(k)$ across concepts. The first panel serves as a baseline: generating from the same model under identical prompts but different random seeds produces a distribution concentrated around $0.5$, confirming that seed variance alone cannot explain the patterns we observe. In most adapted variants, by contrast, we observe heavy-tailed behavior with mass near 0 and 1, indicating systematic concept drift-level drift. This analysis reveals \emph{which} concepts shift, but not whether those shifts translate into a measurable loss of semantic capability. Therefore, we turn to zero-shot classification to answer that question.

\subsubsection{Zero-Shot Degradation.}

If the concept-level shifts surfaced by the SAE analysis are semantically meaningful, they should be reflected in the model's ability to generate recognizable, class-discriminative content. To test this, we evaluate zero-shot classification performance using the Gaussian Diffusion Classifier (GDC)~\cite{qi2024simple}. For each class, GDC generates 240 images and constructs a class prototype by averaging their CLIP image embeddings; test images are then assigned to the nearest prototype by cosine similarity, requiring no real training data at any stage. Classification accuracy therefore depends entirely on the semantic coverage of the generative model: if adaptation erodes a concept, the corresponding prototype degrades and accuracy on that class falls.

We follow the GDC evaluation protocol on three standard benchmarks: CIFAR-10~\cite{krizhevsky2009learning}, Oxford Flowers 102~\cite{nilsback2008automated}, and Food-101~\cite{bossard2014food}. We report both mean accuracy change and worst-case per-class accuracy drop relative to the base model. The gap between these two statistics is the key diagnostic: a small mean drop alongside a large worst-case drop indicates concentrated semantic damage on specific concepts, invisible to aggregate evaluation. Together, the SAE and GDC analyses establish that adaptation inflicts real, concept-specific damage, providing the empirical foundation for \method, which is designed to surface exactly this kind of hidden drift before deployment.

Table~\ref{tab:drift} reports zero-shot classification accuracy changes and distribution-level metrics across all evaluated methods. All FID and KID values are computed between images generated by the base and modified models under identical prompt distributions, isolating the effect of adaptation from any pre-existing gap between the model and its training data. For customization methods, zero-shot results are averaged across models fine-tuned on 10 different concepts; the standard deviation across concepts is therefore a direct measure of how unevenly damage is distributed.

The results reveal three distinct regimes of adaptation damage. The first, illustrated by EraseDiff, is catastrophic and detectable: FID and KID are substantially elevated, and zero-shot accuracy collapses across all three benchmarks, with worst-case drops of $-91.9$ on CIFAR-10 and $-98.1$ on Flowers. Standard aggregate metrics do surface this failure, but at the cost of rendering the model nearly unusable. The second regime, illustrated by SPM, is more insidious: FID and KID remain low (3.643 and 0.00002, respectively), yet zero-shot accuracy drops of up to $-18.9$ on CIFAR-10 reveal that specific concepts have been quietly eroded. The model passes aggregate inspection while silently losing semantic coverage. The third regime, exhibited by DreamBooth across all three base architectures, is the most subtle. FID and KID are consistently low, and \emph{mean} accuracy changes are near zero, yet the standard deviation across concepts tells a different story: on Flowers, SD1.5 shows a mean drop of only $-2.88$ points alongside a standard deviation of $12.21$, implying that while most classes are unaffected, some are severely damaged. No aggregate metric can surface this, because the damage is concentrated and concept-specific. It is precisely the failure mode that \method is designed to expose.

\begin{table*}[t]
\centering
\caption{Worst-case per-class accuracy drop and mean accuracy delta induced by model adaptations, measured via zero-shot classification with GDC. We additionally report FID and KID between images generated by the base and modified models under identical prompt distributions. For each method we report the worst single per-class drop and the mean delta across all classes. For unlearning methods, both metrics are computed on a single target concept (nudity). For DreamBooth, we report mean\,$\pm$\,std across all fine-tuned concepts.}
\label{tab:drift}

\setlength{\tabcolsep}{4pt}
\resizebox{\textwidth}{!}{%
\begin{tabular}{lllcccccc}
\toprule
\textbf{Area} & \textbf{Method} & \textbf{Model} 
& \textbf{FID$\downarrow$} & \textbf{KID$\downarrow$}
& \textbf{CIFAR-10} & \textbf{Flowers-102} & \textbf{Food-101} \\
\midrule
\multirow{4}{*}{Unlearning}
  & AC            & SD1.4 & 10.250 & 0.002 & $-21.5$/$-2.05$ & $-87.3$/$-5.81$ & $-29.5$/$-2.19$ \\
  & EraseDiff     & SD1.4 & 321.621 & 0.377 & $-91.9/-23.24$ & $-98.1$/$-23.60$ & $-87.3$/$-5.68$\\
  & MACE          & SD1.4 & 60.812 & 0.033 & $-39.8/-5.93$ & $-98.1$/$-19.77$ & $-49.1$/$-11.65$ \\
  & SPM           & SD1.4 & 3.643 & 0.00002 & $-18.9/-3.91$ & $-14.6$/$-0.13$ & $-5.0$/$-0.19$ \\
\midrule
\multirow{3}{*}{Customization}
  & DreamBooth    & SD1.5  & $6.86 \pm 2.18$ & $0.002 \pm 0.001$ & $0.96 \pm 7.95$ & $-2.88 \pm 12.21 $ & $0.08 \pm 11.86 $ \\
  & DreamBooth    & SD2.1  & $6.68 \pm 3.66$ & $0.003 \pm 0.003$ & $-2.22 \pm 5.73$ & $ -0.74 \pm 10.91 $ & $-0.91 \pm 12.45$ \\
  & DreamBooth    & SD3.5  & $8.45 \pm 2.33$ & $0.003 \pm 0.001$ & $0.84 \pm 8.88$ & $ -0.21 \pm 10.72 $  & $-1.11 \pm 9.36$ \\
\bottomrule
\end{tabular}%
}
\label{tab:zero_shot}
\vspace{-4mm}
\end{table*}

\subsection{\method: Prompt-Level Model Diffing}

Identifying which concepts are most affected by a model adaptation requires a probe that is both semantically grounded and differentiable; one that can navigate the discrete token space of natural language while remaining sensitive to weight-level changes in a diffusion model. We draw inspiration from two complementary ideas. First, the masked language modeling paradigm~\cite{devlin2018bert}: by treating prompt construction as a fill-mask task, we can optimize \emph{which words} to place in a sentence rather than engineering prompts by hand, letting the optimization surface tokens that are maximally informative about a model's behavior. Second, differentiable prompt optimization methods~\cite{liu2024discovering,carnemolla2025dexter}, which have shown that mapping learnable continuous embeddings to discrete tokens via Gumbel-Softmax relaxation can effectively probe the internal representations of vision and language models. These methods optimize prompts to explain or steer a \emph{single} model; we repurpose the same paradigm for an entirely different goal: \emph{comparing two model checkpoints} to localize where adaptation has caused semantic drift.

\begin{figure}[tb]
    \centering
    \includegraphics[width=0.8\textwidth]{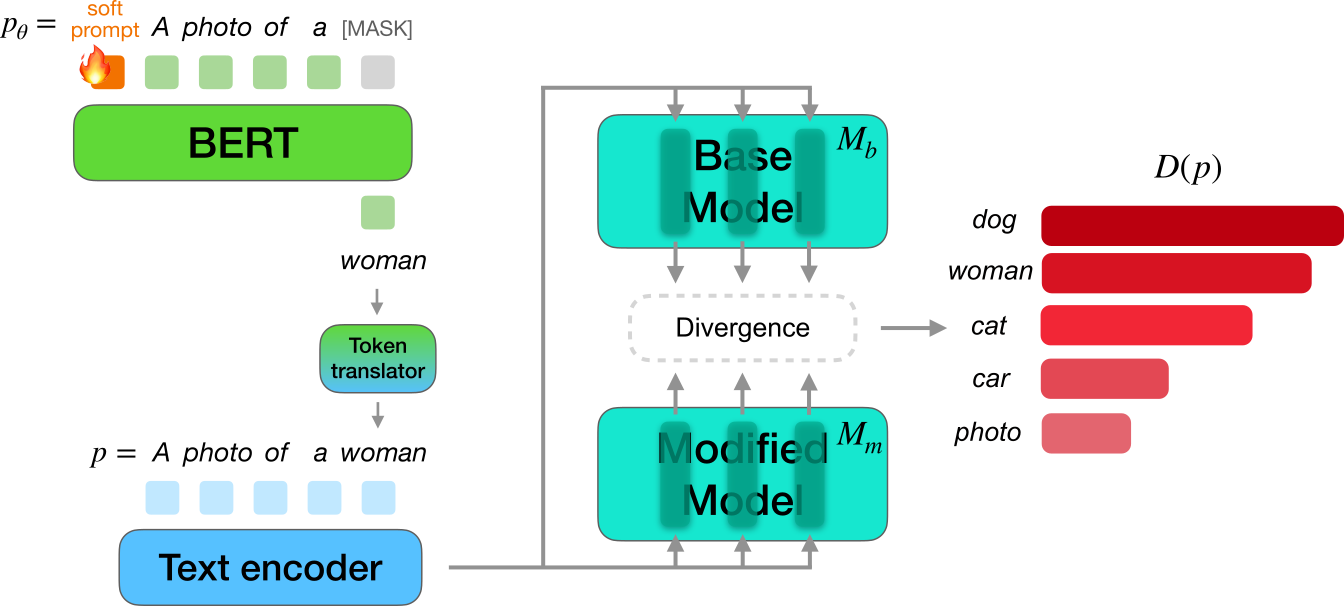}

    \caption{
    \method optimizes a soft prompt $p_\theta = [\mathbf{s}_\theta, \mathbf{t}, \mathbf{m}]$ to find the discrete tokens whose visual concepts diverge most between a base model $\mathcal{M}_b$ and its adapted counterpart $\mathcal{M}_m$. The resulting ranked drift report surfaces concepts most affected by fine-tuning.
    }
    \label{fig:method}
    \vspace{-4mm}
\end{figure}

\method takes a base diffusion model $\mathcal{M}_b$ and its adapted counterpart $\mathcal{M}_m$ and learns a soft prompt whose discrete realization maximizes the divergence between the cross-attention maps produced by each checkpoint under identical noise seeds, depicted in Figure~\ref{fig:method}. Rather than asking ``what does this model focus on?'', we ask ``where do these two models disagree?''. Once the prompt is optimized, divergence is attributed token-by-token, yielding a ranked blacklist of words whose associated visual concepts have shifted most between checkpoints. Because the method is prompt-driven and operates on any pair of checkpoints, it surfaces non-obvious collateral damage (concepts the practitioner never intended to modify) providing an interpretable, concept-level audit before deployment.

\subsubsection{Prompt Parameterization and Encoder Translation.}

We parameterize the input to the masked language model as a sequence $p_\theta = [\mathbf{s}_\theta, \mathbf{t}, \mathbf{m}]$, where $\mathbf{s}_\theta = \{s_i\}$ is a set of learnable soft prompt vectors prepended to the sequence, $\mathbf{t} = \{t_j\}$ is the tokenized fixed text, and $\mathbf{m} = \{m_k\}$ are the mask tokens whose discrete realizations we wish to optimize. The masked language model predicts a distribution over vocabulary entries for each $m_k$; we obtain a differentiable discrete token via Gumbel-Softmax relaxation, enabling gradient-based optimization of $\mathbf{s}$ through standard backpropagation.

Because diffusion architectures employ different text encoders (\eg, CLIP variants in SD 1/2/3/XL, T5 in SD3 and FLUX), the tokens predicted in the masked language model's vocabulary may not map directly into the target encoder's vocabulary. To bridge this gap, we employ a token translator: a fixed binary matrix $M \in \{0,1\}^{V \times W}$ that maps each one-hot token vector from the source vocabulary of size $V$ to its closest counterpart in the target vocabulary of size $W$. This design, used in prior prompt optimization work~\cite{carnemolla2025dexter, shin2020autoprompt}, allows \method to operate across diffusion architectures without modifying the divergence objective or the optimization procedure.

During optimization, both diffusion model checkpoints remain frozen; gradients flow only through the soft prompt parameters $\mathbf{s}_\theta$. To ensure that the tokens discovered generalize beyond a single stochastic realization, we optimize over a batch of $S$ fixed noise seeds rather than a single sample, so that high-scoring tokens reflect consistent cross-model divergence rather than the idiosyncrasies of a particular noise realization.

\subsubsection{\method Objective.}
\label{sec:formal_objective}

We aim to find a prompt that maximally exposes the representational differences induced by adaptation, by measuring how differently the two checkpoints process the same textual input across noise seeds.

\paragraph{Cross-attention extraction.}
For a prompt $p$ and noise seed $z$, both models denoise latent trajectories under the same sampling schedule. At each step, the network produces cross-attention maps that relate textual tokens to spatial latent features. 
Depending on the architecture (UNet-based or transformer-based) layers may operate at varying or identical spatial resolutions. We denote the cross-attention map for a given layer as:

\begin{equation}
A(\mathcal{M}, p, z) \in \mathbb{R}^{H \times W \times T},
\end{equation}
where $H \times W$ is the spatial resolution and $T$ the number of prompt 
tokens. The subset of layers $\mathcal{L}$ most sensitive to adaptation 
drift is architecture-dependent; for SD1.5, for instance, this corresponds to \texttt{down\_block\_0} through \texttt{up\_block\_3}.

\paragraph{Divergence objective.}
We define the divergence for a given prompt as the $L_1$ difference between cross-attention maps, averaged across the seed batch $\{z_1, \dots, z_S\}$ and layers $\mathcal{L}$:

\begin{equation}
D(p) = \frac{1}{S} \sum_{s=1}^{S} \sum_{\ell \in \mathcal{L}}
\left\| A^{(\ell)}(\mathcal{M}_b, p, z_s) - 
A^{(\ell)}(\mathcal{M}_m, p, z_s) \right\|_1.
\end{equation}
The $L_1$ norm was chosen over $L_2$ as it produced sharper token-level 
localization and greater stability across seeds.

\paragraph{Prompt optimization.}
We optimize the soft prompt $\mathbf{s}_\theta$ to maximize $D$:
\begin{equation}
\mathbf{s}_\theta^* = \arg\max_{\mathbf{s}_\theta} \, D(p_\theta),
\end{equation}
where $p_\theta = [\mathbf{s}_\theta, \mathbf{t}, \mathbf{m}]$ follows the parameterization introduced in the prompt parametrization definition. Gradients are backpropagated through both frozen checkpoints simultaneously.

\subsubsection{Word-Level Drift Attribution.}

Once the prompt $p_\theta^* = [\mathbf{s}_\theta^*, \mathbf{t}, \mathbf{m}^*]$ has been optimized, \method assigns a drift score to the discrete tokens realized at the mask positions $\mathbf{m}^*$. Rather than isolating each token's attention slice, the score is computed over the cross-attention maps of the full prompt, capturing how the two checkpoints process the entire scene (including object relations and context) rather than individual concept appearances:
\begin{equation}
\Delta(\mathbf{m}^*) = \frac{1}{S} \sum_{s=1}^{S} \sum_{\ell \in \mathcal{L}} \left\| A^{(\ell)}(\mathcal{M}_b, p^*, z_s) - A^{(\ell)}(\mathcal{M}_m, p^*, z_s) \right\|_1.
\end{equation}
By running the optimization multiple times with different initializations, each run yields a candidate token (or set of tokens) together with its associated drift score $\Delta$. Because scores are aggregated across multiple seeds and layers, the ranking reflects consistent representational change rather than noise from any particular stochastic realization.

\subsubsection{Practical Usage.}
In practice, a user provides a prompt of interest and two checkpoints; \method returns a ranked list of tokens whose visual concepts exhibit the 
largest cross-model divergence, or effectively, a blacklist of concepts that have been most affected by adaptation. By inverting the objective, \method can equally identify tokens that minimize $\Delta$, revealing concepts that remain stable across checkpoints. For deployment scenarios, practitioners can run \method over a set of anticipated prompts and aggregate the most frequently appearing high-drift tokens, surfacing systematically unstable concepts before an adapted model is released.

\section{Experiments}
\label{sec:experiments}

\subsection{Experimental setup}
\subsubsection{Evaluation Protocol.}

We evaluate \method by measuring whether the tokens it identifies as high-drift produce maximally divergent image distributions between the base and modified models.

\paragraph{Template Prompts.}
We choose five template prompts from Textual Inversion~\cite{gal2023textualinversion} formulations (\eg, ``A photo of a \texttt{[MASK]}''). These templates provide controlled semantic contexts while allowing \method to optimize the target concept token (see Supplementary Material for details).

\paragraph{Drift Optimization.}
For each template, we run the \method optimization procedure 100 times with different prompt initialization seeds, yielding 500 optimization runs per method. To reduce memory requirements and wall-clock time, we restrict denoising to the first 4 steps of the diffusion trajectory; early steps govern high-level semantic structure and are most sensitive to concept-level drift~\cite{biroli2024dynamical}, making further steps unnecessary for reliable drift signal. In each run, we extract the top-ranked drift tokens according to the word-level divergence score defined in Section~\ref{sec:formal_objective}. We then aggregate results across runs and select the five most frequently occurring high-drift tokens per method.


\paragraph{Image Generation.}
For each selected token, we generate 100 images from each checkpoint under identical sampling conditions (seeds, resolution, guidance scale, and number of denoising steps), ensuring that observed differences arise solely from the checkpoint weights. DreamBooth models are fine-tuned on the publicly available dog images provided by the DreamBooth authors (shown in Supplementary Material). Unlearning checkpoints are taken from the AdvUnlearn repository\footnote{\url{https://github.com/OPTML-Group/AdvUnlearn?tab=readme-ov-file\#checkpoints}}.

\paragraph{Distribution Comparison.}
This procedure produces paired image sets for each high-drift token. We quantify divergence between the base and modified image distributions using the metrics described below. All experiments are carried on identical prompt distributions and sampling configurations across methods.

\subsubsection{Metrics.}

We quantify divergence between image sets generated by the base and modified models using four complementary metrics. CLIP-I measures the average cosine similarity between CLIP image embeddings of corresponding pairs generated under identical prompts and noise seeds — lower values indicate greater semantic divergence. MS-SWD~\cite{he2024ms-swd} captures discrepancies in multi-scale image statistics, including structural and textural differences, with higher values indicating greater distributional divergence. LPIPS~\cite{zhang2018lpips} measures perceptual distance using deep feature activations from a pretrained network, correlating more closely with human judgments than pixel-wise metrics, with higher values indicating greater perceptual divergence. Finally, Q-Eval~\cite{zhang2025q} evaluates prompt-image alignment and perceptual quality using a unified metric trained on large-scale human preference annotations; a drop in score indicates that adaptation has degraded the model's ability to generate a concept both perceptually and semantically.

\subsection{Quantitative Results}

We evaluate \method on a representative subset spanning different damage regimes: ESD~\cite{gandikota2023erasing} and SPM~\cite{lyu2023onedimensional} for concept unlearning, Scissorhands (SH)~\cite{wu2024scissorhands} as a degenerate-regime baseline, and DreamBooth across three base architectures for concept customization. For each method, we evaluate across 6 concepts spanning diverse semantic categories. \method identifies the five most frequently occurring high-drift tokens across 500 optimization runs per concept.


Tables~\ref{table:max_drift} and~\ref{table:min_drift} jointly structure the evaluation: Table~\ref{table:max_drift} reports metrics for high-drift tokens surfaced by \method, while Table~\ref{table:min_drift} serves as an internal reference for low-drift tokens identified by inverting the objective. If \method is working, the gap between the two tables should be consistent and interpretable. The ranked token lists for each method and concept are provided in Appendix~\ref{app:top_words_full}.

The clearest illustration comes from SPM. High-drift tokens yield a CLIP-i of 0.88 and LPIPS of 0.33, while low-drift tokens reach 0.96 and 0.14 respectively, a sharp separation that confirms \method is reliably discriminating between stable and unstable concepts. The words themselves reinforce the paper's central claim, as unlearning nudity surfaces \textit{body, goddess, woman} as collaterally damaged concepts that the practitioner never intended to modify\footnote{The high-drift tokens can be found in the Supplementary material Table 4}.

ESD presents a more complex picture. The high-drift vocabulary (\textit{seal, cross, boyfriend}) is semantically distant from the nudity target, illustrating non-obvious collateral damage, yet the gap between high- and low-drift conditions is modest across metrics, suggesting ESD induces broader, less localized damage that makes clean separation harder.

Scissorhands represents the degenerate regime described in Section~\ref{sec:method}: a Q-Eval of 0.12 and CLIP-i of 0.56 on high-drift tokens indicate widespread model degradation rather than targeted concept erosion. As discussed in the Limitations, \method's attributions for this method should be interpreted with caution.

For DreamBooth, the most informative axis is Q-Eval, which captures both perceptual quality and prompt faithfulness. As shown in Table~\ref{tab:drift}, mean accuracy changes are near zero across all three architectures, yet the standard deviation reveals concept-specific instability. Accordingly, SD2.1 drops to 0.39 on high-drift tokens while SD3.5 remains at 0.69, suggesting the latter is more robust to collateral damage. The drift vocabulary is also telling: SD3.5 surfaces \textit{bible, dragon, devil} as the concepts most affected by a dog fine-tune, while SD2.1 includes \textit{dog} itself alongside \textit{woman} and \textit{car}, all examples of non-obvious collateral damage on concepts the practitioner never intended to modify.

A key property of \method is that the tokens it surfaces should reflect the specific concept being adapted rather than a generic fine-tuning signature. The results confirm that collateral damage is structured and target-dependent: the drift vocabulary shifts predictably with the adapted concept, tracing damage back to its source in the model's representational space. Crucially, this does not mean the damaged concepts are obvious or intended — unlearning \textit{garbage} displaces \textit{truck} and \textit{bus}, while unlearning \textit{tench} affects \textit{book} and \textit{calendar}, none of which a practitioner would anticipate. This target-dependence is precisely what makes \method actionable: it does not return a generic blacklist, but a drift report that is specific to each adaptation.

\begin{table}[]
\vspace{-4mm}
\caption{High-drift tokens and image-level divergence metrics identified by \method.}
\label{table:max_drift}
\setlength{\tabcolsep}{4pt}
\resizebox{\textwidth}{!}{%
\begin{tabular}{lcccccl}
\hline
\textbf{Method} & \textbf{Model} & \textbf{CLIP-i} $\uparrow$& \makecell{\textbf{MS-SWD}$\downarrow$} & \makecell{\textbf{LPIPS}$\downarrow$}& \makecell{\textbf{Q-Eval}$\uparrow$}\\
\hline
\multicolumn{6}{c}{Unlearning}                                                        \\
\hline
ESD     &   SD1.4   & 0.72 {\scriptsize $\pm $ 0.13} &  1.84   & 0.57 {\scriptsize $\pm $ 0.15}  & 0.41 {\scriptsize $\pm $ 0.17} \\
SPM    &   SD1.4    &    0.88 {\scriptsize $\pm $ 0.11}    &   1.02 & 0.33 {\scriptsize $\pm $ 0.20} & 0.50 {\scriptsize $\pm $ 0.16}
 \\
SH    &   SD1.4    &    0.56 {\scriptsize $\pm $ 0.09}    &  4.15  & 0.82 {\scriptsize $\pm $ 0.08}   & 0.12 {\scriptsize $\pm $ 0.07} \\
\hline
\multicolumn{6}{c}{Customization}                                                     \\
\hline
Dreambooth     &   SD1.5    &    0.81  {\scriptsize $\pm $ 0.13}   &  1.87  & 0.51 {\scriptsize $\pm $ 0.18}  & 0.52 {\scriptsize $\pm $ 0.18}\\
Dreambooth     &   SD2.1    &    0.81 {\scriptsize $\pm $ 0.10}    &  1.81  & 0.54 {\scriptsize $\pm $ 0.14} & 0.39 {\scriptsize $\pm $ 0.15}\\
Dreambooth     &   SD3.5    &    0.81  {\scriptsize $\pm $ 0.10}  &  2.20  & 0.54 {\scriptsize $\pm $ 0.13}  & 0.69 {\scriptsize $\pm $ 0.17} \\
\hline
\end{tabular}
}
\end{table}

\begin{table}[]
\caption{ Low-drift tokens and image-level divergence metrics identified by \method.}
\label{table:min_drift}
\setlength{\tabcolsep}{4pt}
\resizebox{\textwidth}{!}{%
\begin{tabular}{lcccccl}
\hline
\textbf{Method} & \textbf{Model} & \textbf{CLIP-i} $\uparrow$& \makecell{\textbf{MS-SWD}$\downarrow$} & \makecell{\textbf{LPIPS}$\downarrow$} & \makecell{\textbf{Q-Eval}$\uparrow$}&\\
\hline
\multicolumn{6}{c}{Unlearning}                                                        \\
\hline

ESD   &   SD1.4   & 0.79 {\scriptsize $\pm $ 0.11}   &  1.74   & 0.52  {\scriptsize $\pm $ 0.13} & 0.41 {\scriptsize $\pm $ 0.12} & \\
SPM    &   SD1.4    &    0.96 {\scriptsize $\pm $ 0.05}    &   0.39 & 0.14 {\scriptsize $\pm $ 0.11}  & 0.43 {\scriptsize $\pm $ 0.14} &
 \\
SH    &   SD1.4    &   0.62  {\scriptsize $\pm $ 0.10}   &  4.01 & 0.78 {\scriptsize $\pm $ 0.11}  & 0.19 {\scriptsize $\pm $ 0.08}  &  \\
\hline
\multicolumn{6}{c}{Customization}                                                     \\
\hline
Dreambooth     &   SD1.5    &    0.85 {\scriptsize $\pm $ 0.10}    &  1.41  & 0.44  {\scriptsize $\pm $ 0.16} & 0.56 {\scriptsize $\pm $ 0.16} \\
Dreambooth     &   SD2.1    &    0.84 {\scriptsize $\pm $ 0.09}     &  1.44  & 0.50 {\scriptsize $\pm $ 0.15} & 0.42 {\scriptsize $\pm $ 0.16} \\
Dreambooth     &   SD3.5    &    0.81 {\scriptsize $\pm $ 0.10}   &  2.11  & 0.53 {\scriptsize $\pm $ 0.13}  & 0.71 {\scriptsize $\pm $ 0.16} \\
\hline

\end{tabular}
}
\end{table}

\subsection{Qualitative Results}

\begin{figure}[htbp]
    \vspace{-3mm}
    \centering
    \begin{subfigure}[b]{0.49\textwidth}
        \centering
        \includegraphics[width=\textwidth]{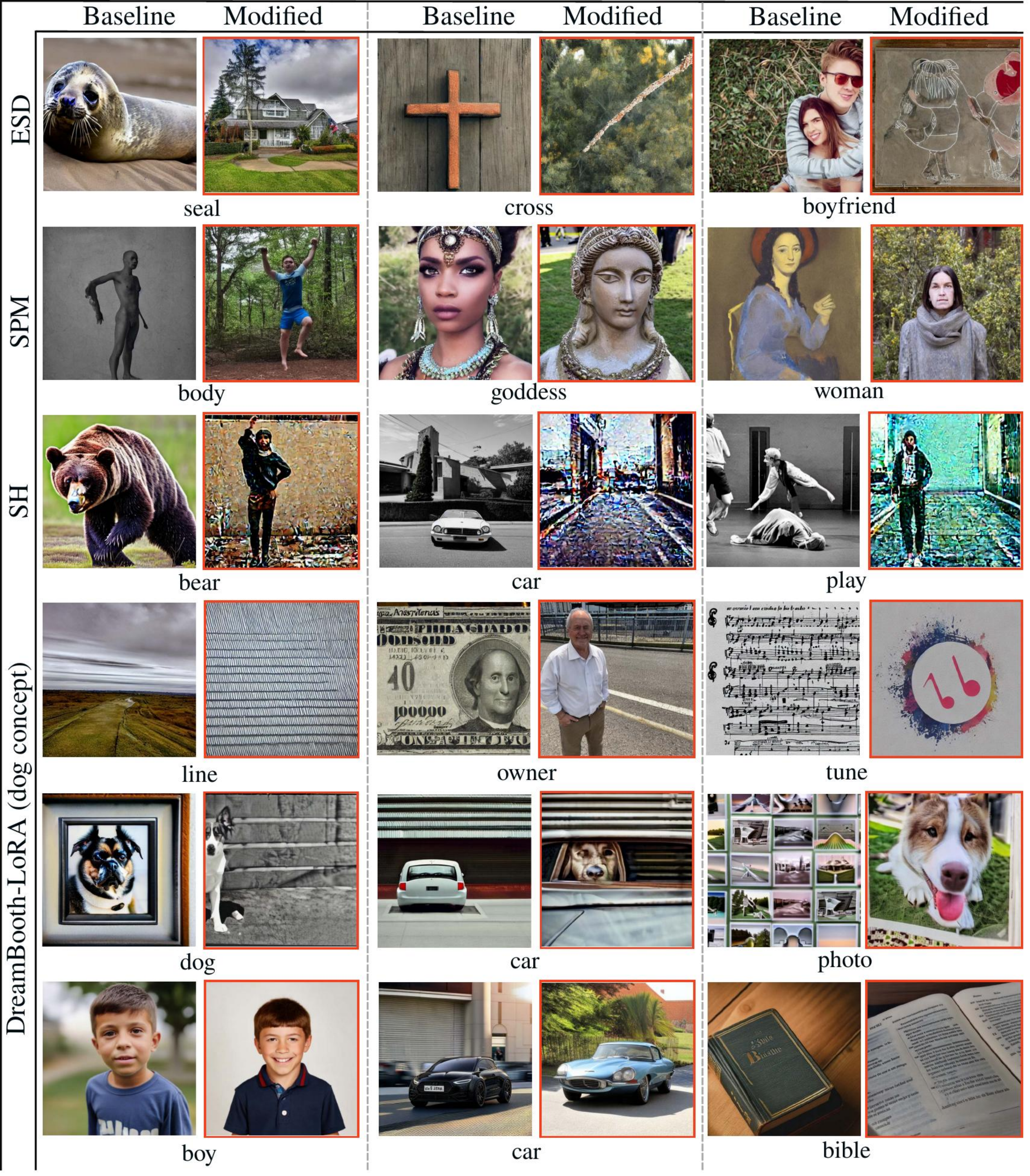}
        \caption{\textbf{Maximized drift.}}
        \label{fig:max_drift_left}
    \end{subfigure}
    \hfill
    \begin{subfigure}[b]{0.49\textwidth}
        \centering
        \includegraphics[width=\textwidth]{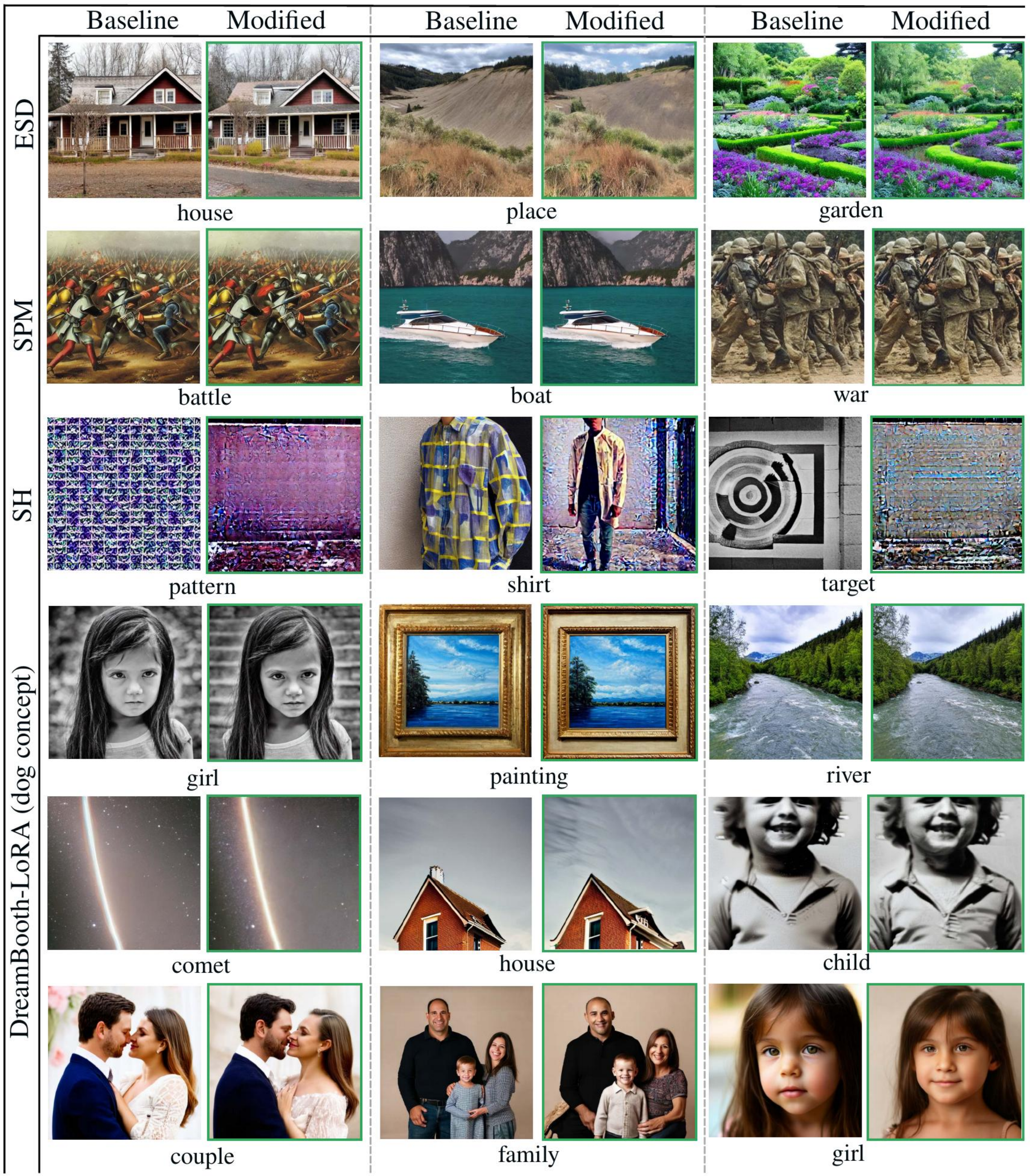}
        \caption{\textbf{Minimized drift.}}
        \label{fig:max_drift_right}
    \end{subfigure}
    \caption{Qualitative examples of prompts found by DriftScope that (a) \emph{maximize} and (b) \emph{minimize} cross-attention drift between model checkpoints. We report results for unlearning methods (ESD, SPM, and Scissorhands) and DreamBooth LoRA customization (dog concept) across Stable Diffusion 1.5, 2.1, and 3.5-Medium.}
    \label{fig:max_drift_total}
\end{figure}

Figure~\ref{fig:max_drift_total} provides the visual counterpart to the quantitative results. High-drift pairs show changes well beyond stylistic variation: for ESD, \textit{seal} shifts from a marine animal to a house exterior and \textit{boyfriend} transforms from a portrait to a sketch; for SPM, \textit{body} and \textit{goddess} both shift markedly in composition and style, suggesting nudity unlearning broadly reshapes human-form rendering. In DreamBooth, fine-tuning bleeds into adjacent tokens such as \textit{bible} and \textit{car} in SD3.5 and \textit{dog} itself in SD2.1. Low-drift tokens (SPM's \textit{battle}, \textit{boat}; DreamBooth's \textit{girl}, \textit{painting}) produce near-identical pairs, confirming \method discriminates between stable and unstable concepts. Scissorhands outputs remain degraded in both conditions, consistent with the widespread model collapse identified quantitatively.

\section{Ablations}

\begin{figure}[htbp]
     \vspace{-4mm}
    \centering
    \includegraphics[width=\textwidth]{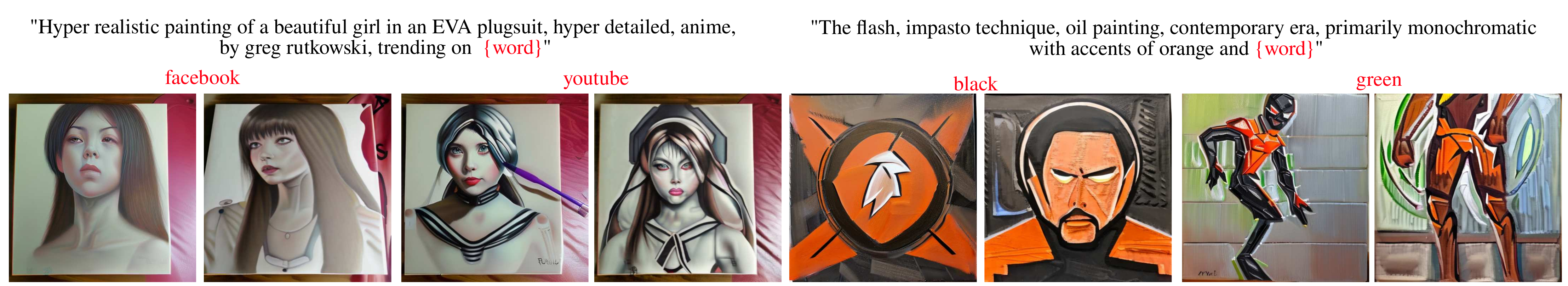}
        \caption{\method generalizes beyond short single-concept prompts. Shown are high-drift image pairs from SD~2.1 under long, naturalistic inputs sourced from PromptHero\cite{prompthero}, for a DreamBooth model fine-tuned on a dog concept. Each pair shows the base model generation (left) alongside the adapted model generation (right).}
        \label{fig:large_prompts}
\end{figure}

\paragraph{Generalization to Long and Domain-Specific Prompts.}
\method is not limited to short, single-concept prompts. Figure~\ref{fig:large_prompts} shows that collateral damage surfaces equally in longer, naturalistic inputs: in the DreamBooth dog fine-tune, \textit{Facebook} emerges as a high-drift token despite having no relation to the adapted concept, and color-bearing prompts show stylistic drift propagating to modifiers the practitioner would never expect to be affected. This confirms that \method's utility extends to the open-ended prompts encountered in real deployment.

\section{Limitations}
\method is optimization-based rather than analytic, searching across many initializations and averaging over multiple seeds, which makes it computationally intensive while mitigating sensitivity to degenerate noise realizations. The method also assumes both checkpoints retain meaningful generative structure: when adaptation causes widespread degradation (as with Scissorhands), cross-attention maps become globally uninformative and concept-level attribution breaks down.

\section{Conclusion}
\label{sec:conclusion}

Adapting text-to-image diffusion models is routinely treated as a benign operation, evaluated only on its intended effects. We have shown this framing is insufficient. Across concept customization and concept unlearning weight-level modification systematically damages semantically unrelated concepts in ways that aggregate metrics structurally cannot surface. When the damage is severe enough for FID and KID to respond, the model is already nearly unusable; when the model remains functional, FID and KID stay flat while specific classes silently suffer worst-case zero-shot accuracy drops of up to 18.9 points and concept-level distributions shift dramatically. 

Building on this evidence, we introduced \method, a prompt-level diagnostic that identifies which words have their visual concepts most affected by adaptation. By maximizing divergence between the cross-attention maps of two checkpoints, \method surfaces non-obvious collateral damage and returns a ranked blacklist of high-drift tokens for any user-provided prompt, without requiring access to real data or model internals beyond standard attention outputs.

\section*{Acknowledgments}
 This work was supported by Grants PID2022-143257NB-I00, PID2024-162555OB-I00, and AIA2025-163919-C52  funded by MICIU/AEI/10.13039/501100011033, \\ERDF/EU and the FEDER, by the Generalitat de Catalunya CERCA Program, by the grant Càtedra ENIA UAB-Cruïlla (TSI-100929-2023-
2) from the Ministry of Economic Affairs and Digital Transition of Spain, by the European Union’s Horizon Europe research and innovation programme under grant agreement number 101214398 (ELLIOT), and by the BBVA Foundations of Science program on Mathematics, Statistics, Computational Sciences and Artificial Intelligence (grant VIS4NN). 
Kai Wang acknowledges the funding No.R002026G0116 and No.R002026B0121 from Guangdong and Hong Kong Universities 1+1+1 Joint Research Collaboration Scheme, and the start-up grant B01040000108 from CityU-DG.
Yiping Han acknowledges the Chinese Scholarship Council (CSC) grant No.202504910071.
We acknowledge the EuroHPC Joint Undertaking for awarding us access to Leonardo at CINECA, Italy, and the RES resources provided by BSC on MareNostrum5 under projects IM-2025-3-0025 and IM-2025-3-0027.

%

\newpage

\bibliographystyle{splncs04}
\bibliography{main}

\newpage
\appendix

\section{Full SAE Drift Score Distributions}
\label{app:blindspots_full}

\begin{figure}[p]
    \centering

    \includegraphics[width=0.24\textwidth]{figures/blindspots_hist/baseline_sd14_seed_diff.pdf}\hfill
    \phantom{\includegraphics[width=0.24\textwidth]{figures/blindspots_hist/baseline_sd14_seed_diff.pdf}}\hfill
    \phantom{\includegraphics[width=0.24\textwidth]{figures/blindspots_hist/baseline_sd14_seed_diff.pdf}}\hfill
    \phantom{\includegraphics[width=0.24\textwidth]{figures/blindspots_hist/baseline_sd14_seed_diff.pdf}}

    \smallskip
    \noindent\textit{Baseline (seed variance control)}
    \vspace{0.5em}

    \includegraphics[width=0.24\textwidth]{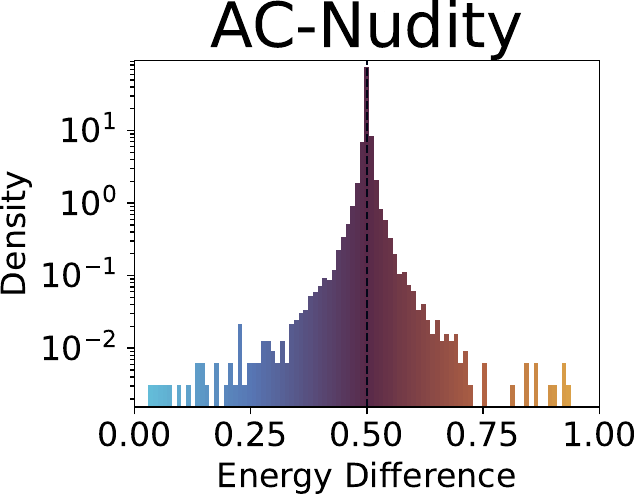}\hfill
    \includegraphics[width=0.24\textwidth]{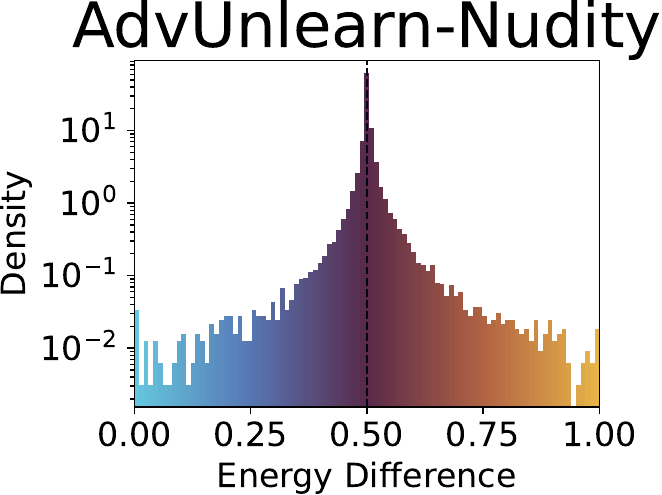}\hfill
    \includegraphics[width=0.24\textwidth]{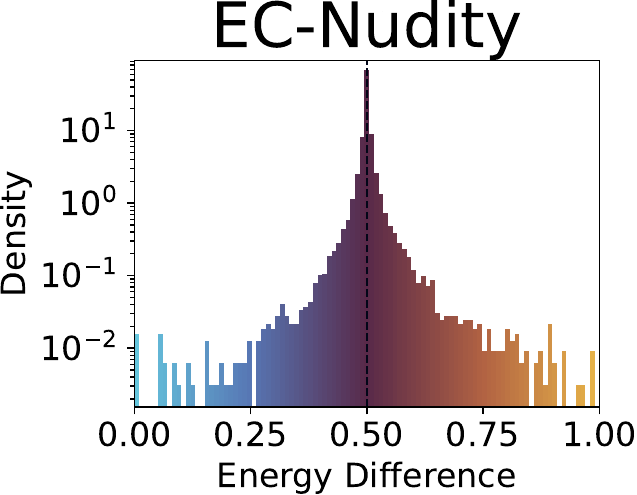}\hfill
    \includegraphics[width=0.24\textwidth]{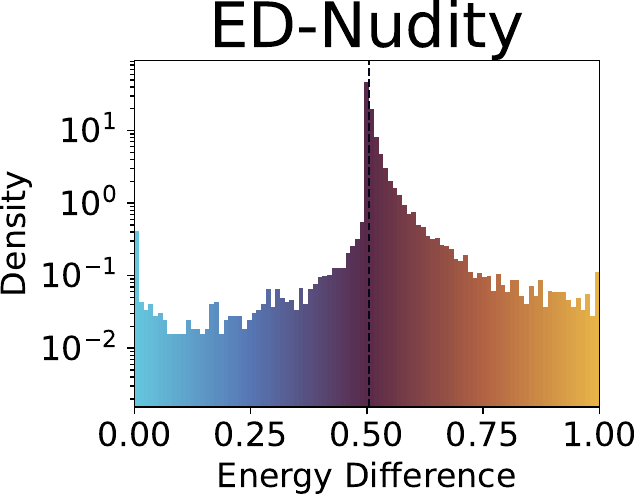}

    \vspace{0.5em}

    \includegraphics[width=0.24\textwidth]{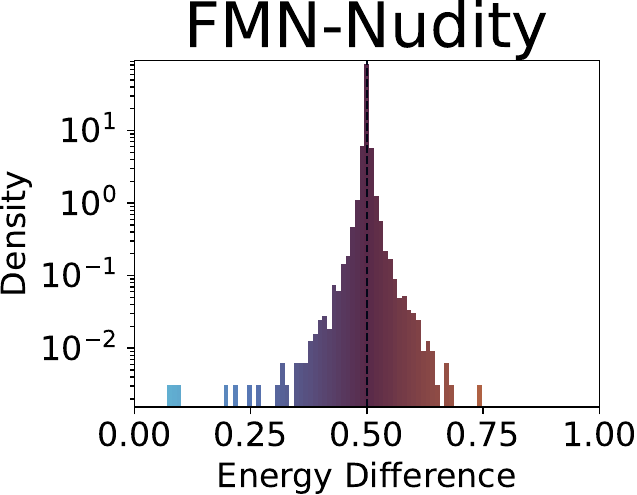}\hfill
    \includegraphics[width=0.24\textwidth]{figures/blindspots_hist/mace_nudity.pdf}\hfill
    \includegraphics[width=0.24\textwidth]{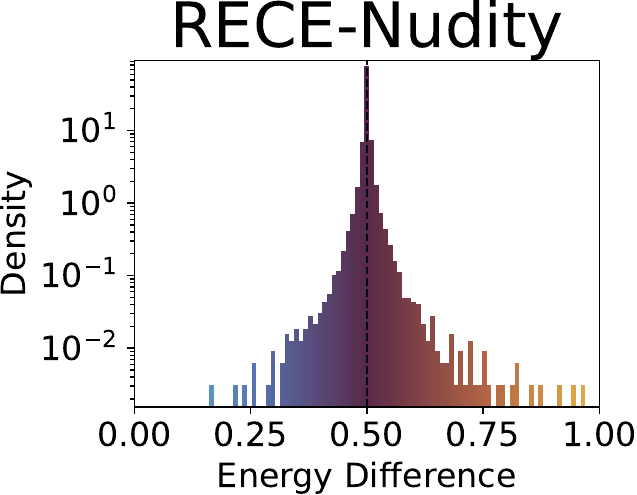}\hfill
    \includegraphics[width=0.24\textwidth]{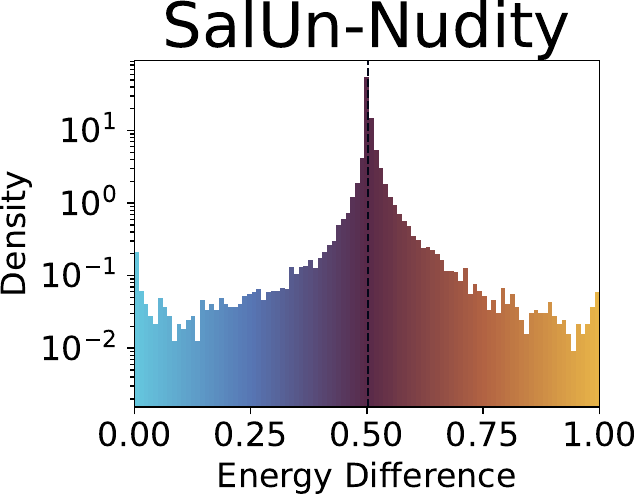}

    \vspace{0.5em}

    \includegraphics[width=0.24\textwidth]{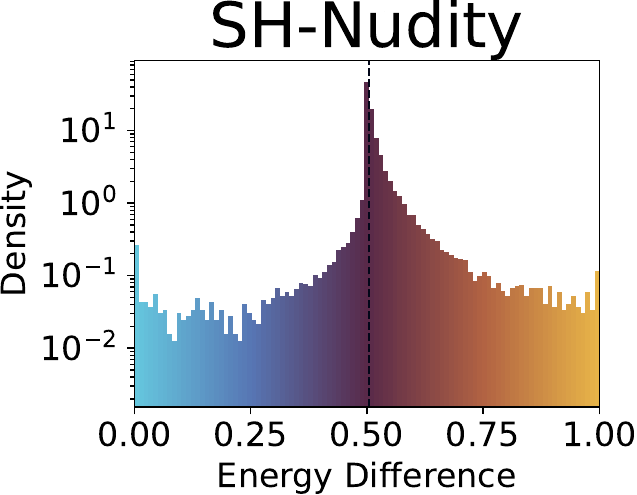}\hfill
    \includegraphics[width=0.24\textwidth]{figures/blindspots_hist/spm_nudity.pdf}\hfill
    \includegraphics[width=0.24\textwidth]{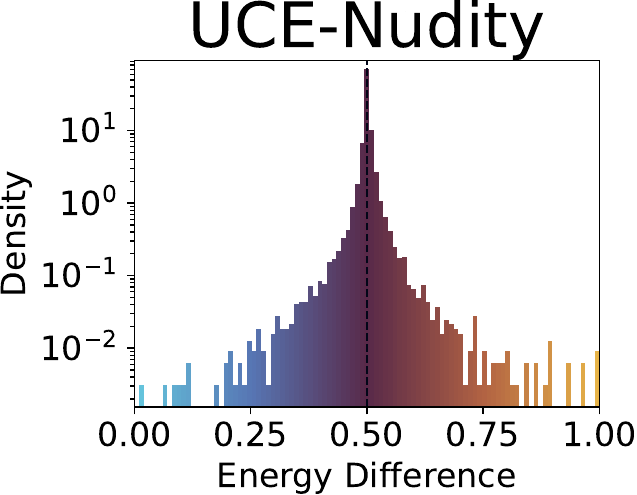}\hfill
    \phantom{\includegraphics[width=0.24\textwidth]{figures/blindspots_hist/baseline_sd14_seed_diff.pdf}}

    \smallskip
    \noindent\textit{Concept unlearning (nudity target): AC, AdvUnlearn, EC, EraseDiff, FMN, MACE, RECE, SalUN, Scissorhands, SPM, UCE}
    \vspace{0.5em}

    \includegraphics[width=0.24\textwidth]{figures/blindspots_hist/boft_dog.pdf}\hfill
    \includegraphics[width=0.24\textwidth]{figures/blindspots_hist/cd_lora_dog.pdf}\hfill
    \includegraphics[width=0.24\textwidth]{figures/blindspots_hist/db_dog.pdf}\hfill
    \includegraphics[width=0.24\textwidth]{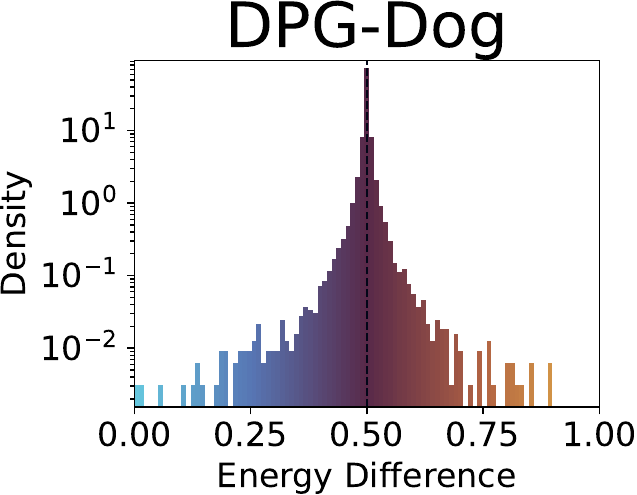}

    \vspace{0.5em}

    \includegraphics[width=0.24\textwidth]{figures/blindspots_hist/svdiff_dog.pdf}\hfill
    \includegraphics[width=0.24\textwidth]{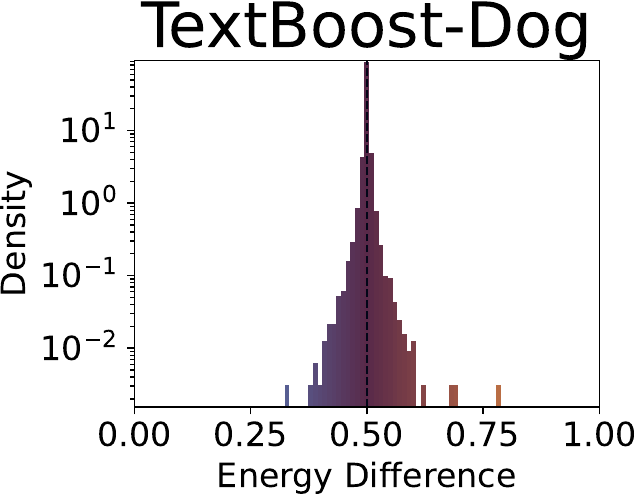}\hfill
    \phantom{\includegraphics[width=0.24\textwidth]{figures/blindspots_hist/baseline_sd14_seed_diff.pdf}}\hfill
    \phantom{\includegraphics[width=0.24\textwidth]{figures/blindspots_hist/baseline_sd14_seed_diff.pdf}}

    \smallskip
    \noindent\textit{Concept customization (dog concept): BOFT, CD-LoRA, DreamBooth, DPG, SVDiff, TextBoost}





    \caption{Full distribution of SAE drift scores $\omega(k)$ across all evaluated methods and paradigms. The baseline panel (top-left) is reproduced from Figure~\ref{fig:grid_blindspots} for reference. Within each paradigm, panels are ordered alphabetically by method name. Heavy-tailed behavior near 0 and 1 indicates systematic concept-level drift; the degree and shape of the tails vary across methods but the pattern is consistent within each paradigm.}
    \label{fig:grid_blindspots_full}
\end{figure}

Figure~\ref{fig:grid_blindspots_full} extends Figure~\ref{fig:grid_blindspots} to all evaluated adaptation paradigms and methods. The baseline panel (top-left) is reproduced for reference. All remaining panels show adapted variants grouped by paradigm: concept unlearning (nudity target) and concept customization (dog concept). In all cases the baseline concentrates sharply around $\omega(k)=0.5$, while adapted variants exhibit heavy-tailed behavior to varying degrees, confirming that the phenomenon is not specific to any single method or paradigm.

\section{DreamBooth Fine-Tuning Data}
\label{app:dreambooth_data}

\begin{figure}[h]
    \centering
    \includegraphics[width=0.18\textwidth]{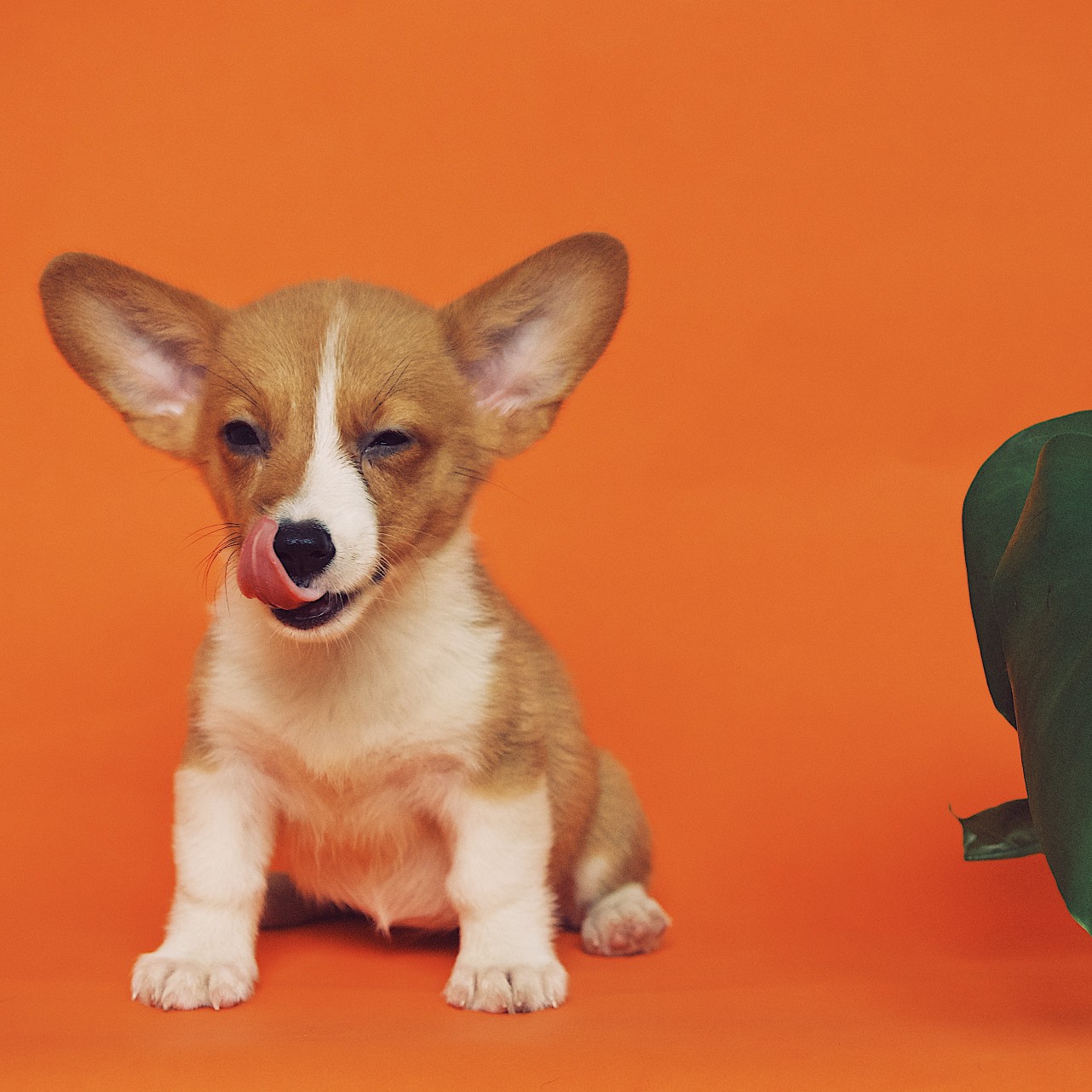}\hfill
    \includegraphics[width=0.18\textwidth]{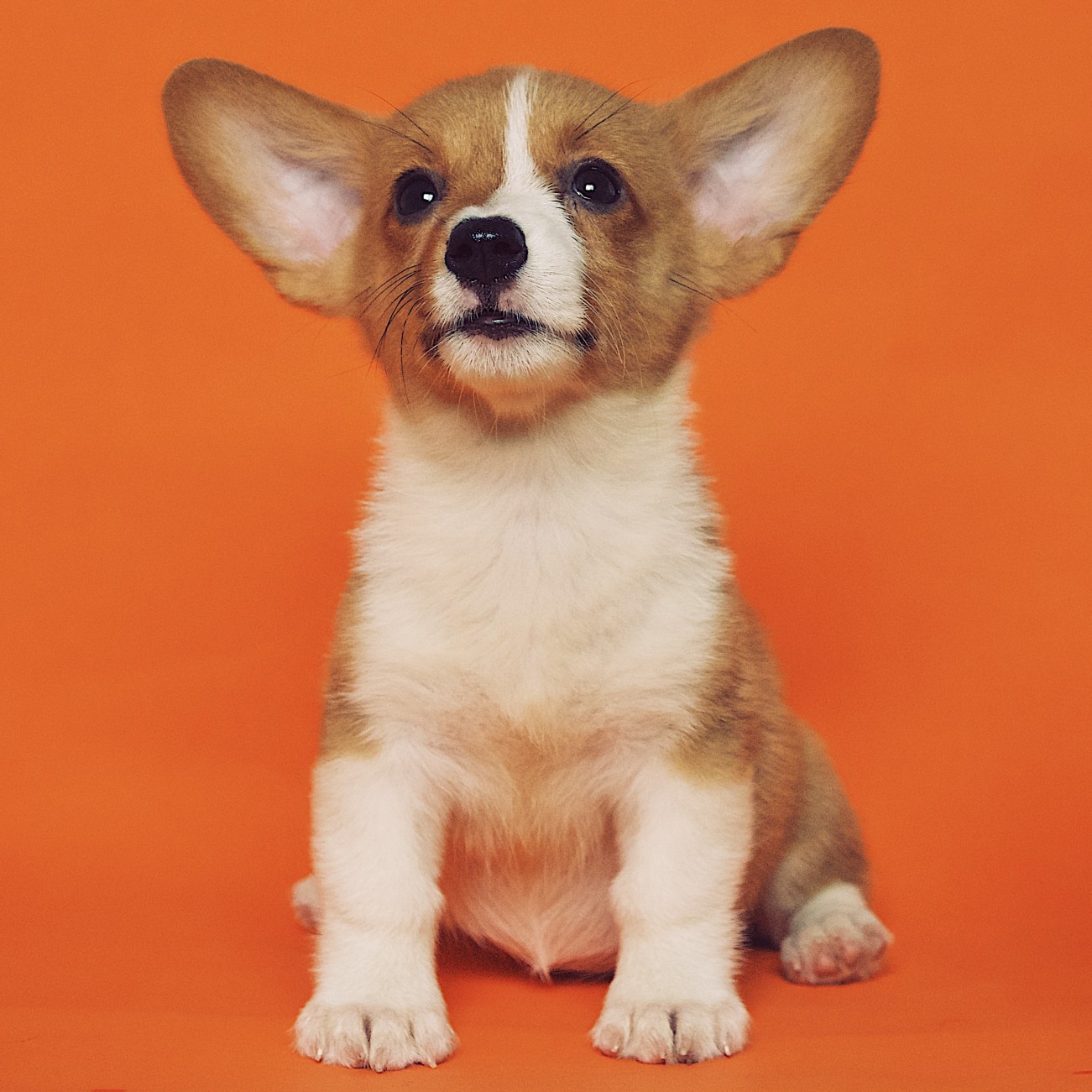}\hfill
    \includegraphics[width=0.18\textwidth]{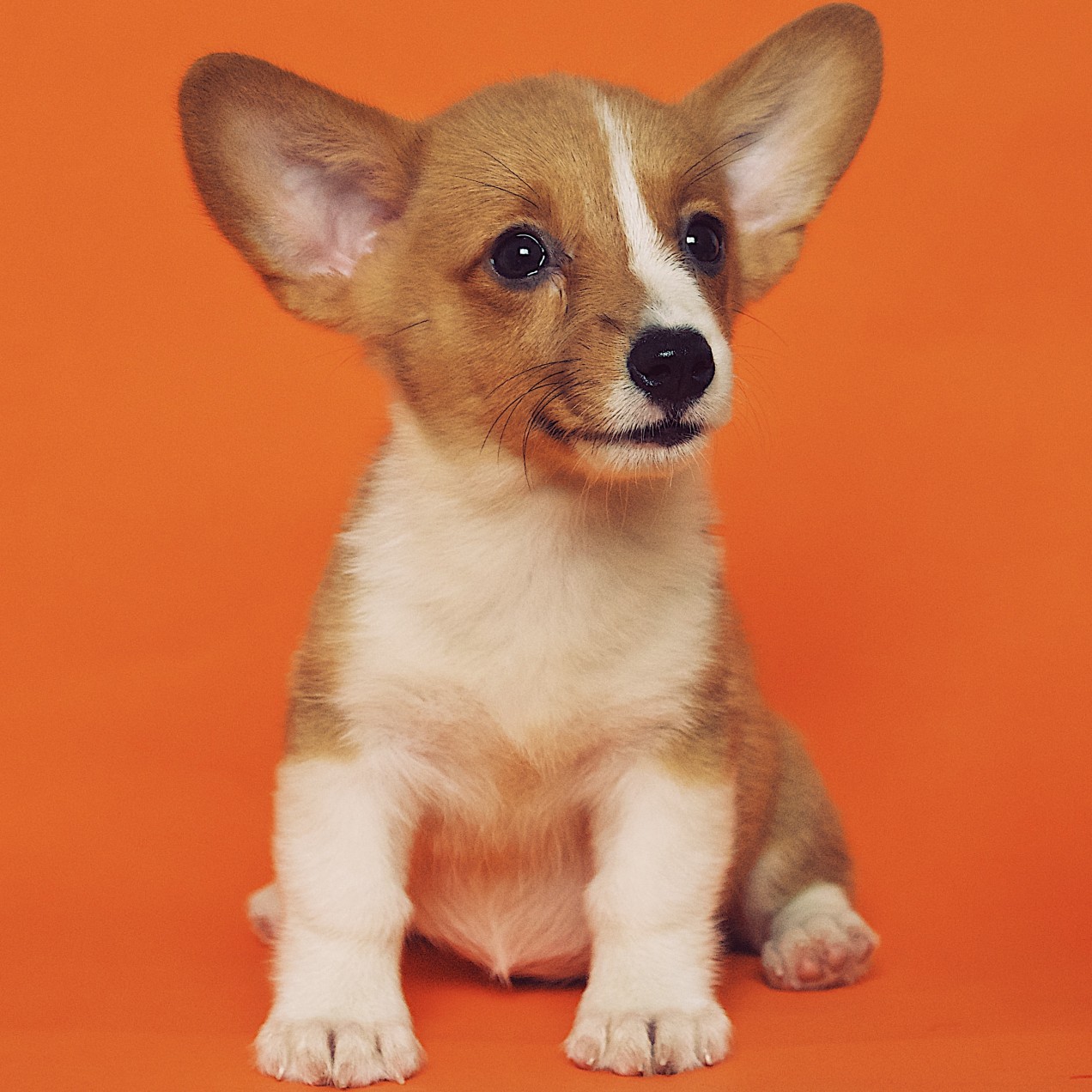}\hfill
    \includegraphics[width=0.18\textwidth]{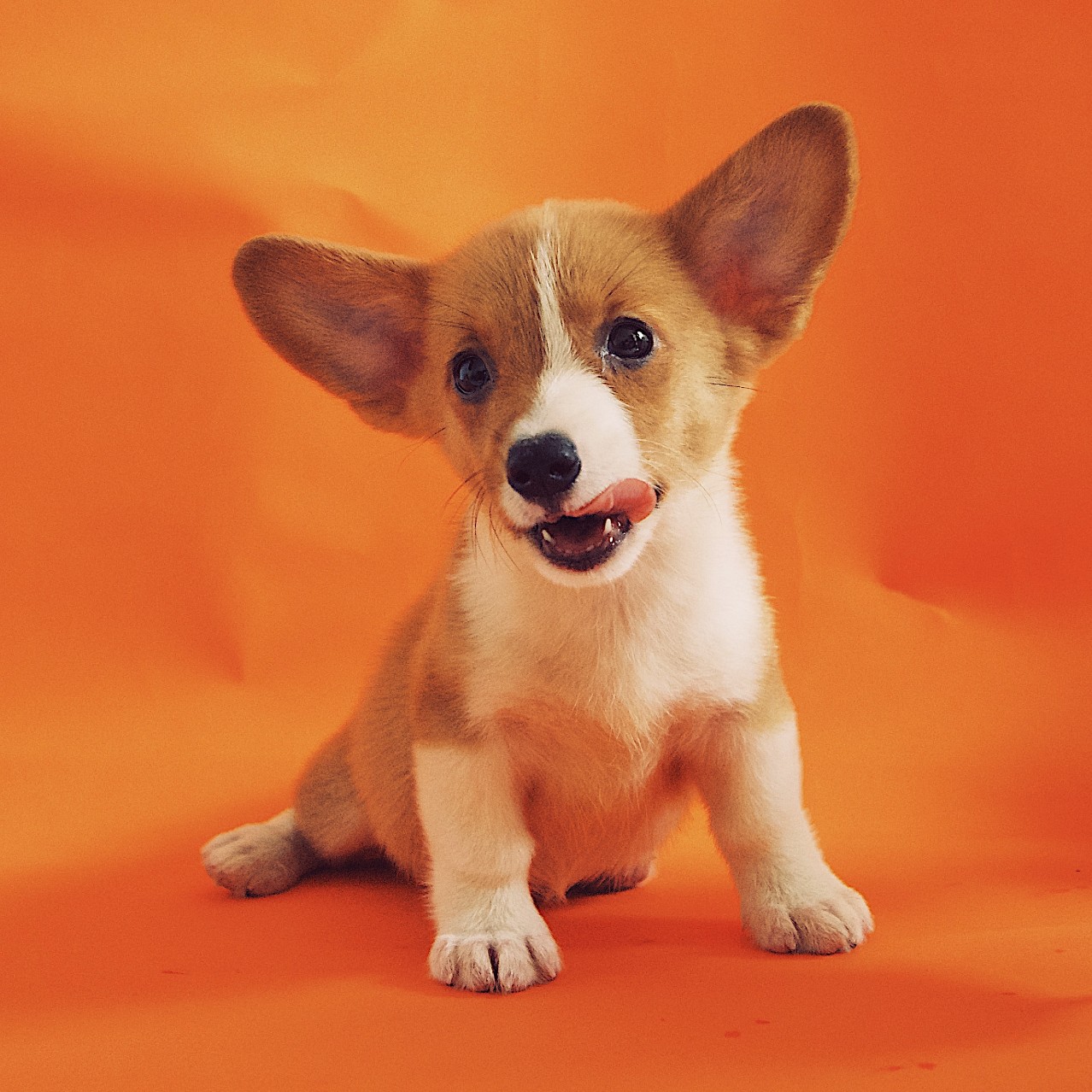}\hfill
    \includegraphics[width=0.18\textwidth]{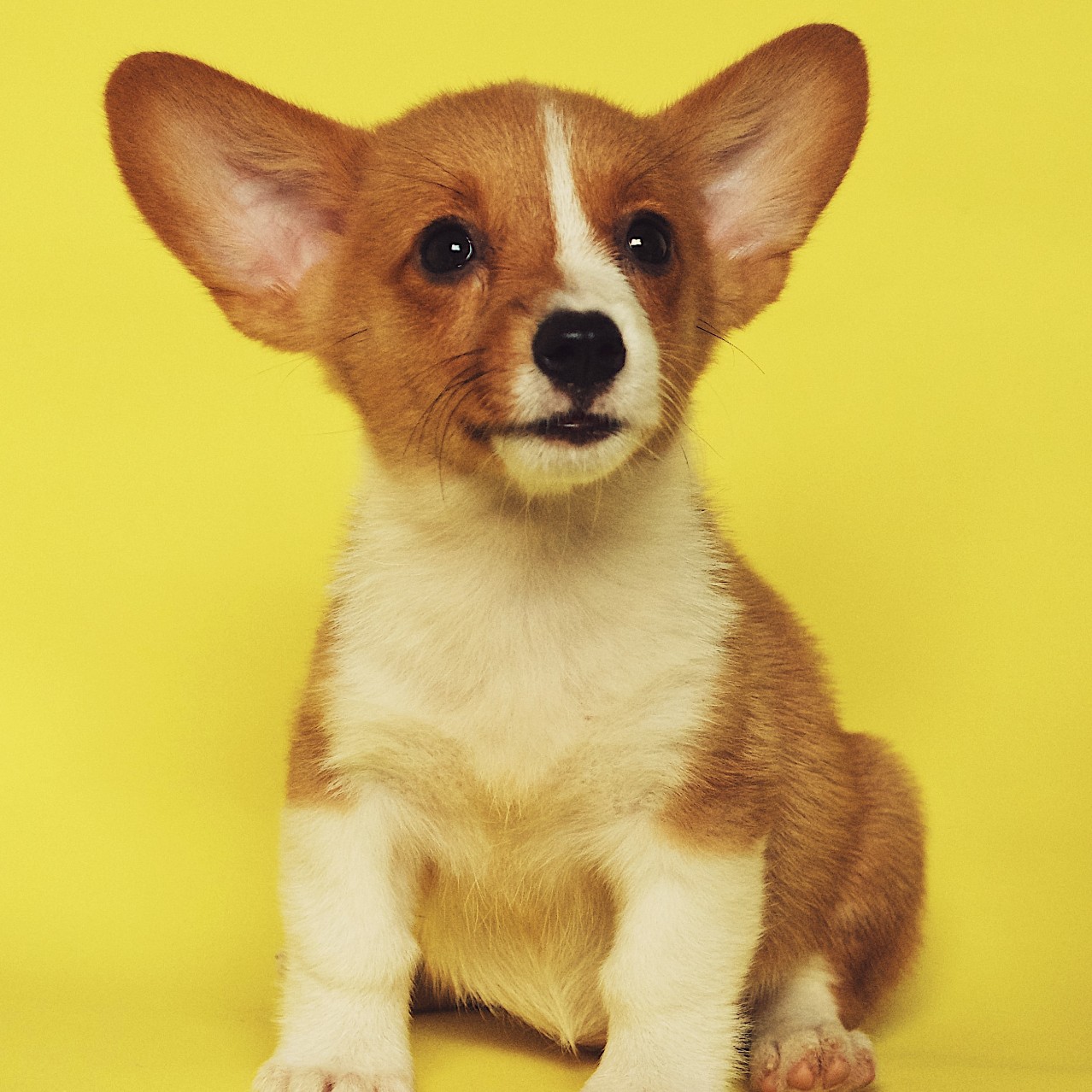}
    \caption{The five dog images used for DreamBooth fine-tuning, sourced from the publicly available dataset released by~\cite{ruiz2023dreambooth}.}
    \label{fig:dreambooth_data}
\end{figure}

Figure~\ref{fig:dreambooth_data} shows the five images used to fine-tune the DreamBooth models evaluated in this paper, taken from the publicly available dog dataset provided by the DreamBooth authors~\cite{ruiz2023dreambooth}.

\section{Template Prompts}
Below we provide the template prompts chosen from Textual Inversion~\cite{gal2023textualinversion}:
\begin{itemize}
\item[$\bullet$] A photo of a [MASK]
\item[$\bullet$] A rendition of the [MASK]
\item[$\bullet$] A photo of a dirty [MASK]
\item[$\bullet$] A rendering of a [MASK]
\item[$\bullet$] A dark photo of a [MASK]
\end{itemize}

\section{Top words for each concept} \label{app:top_words_full}

\begin{table}[t]
\caption{High-drift tokens identified by \method on models adapted to unlearn different concepts. Each row corresponds to an independently adapted checkpoint.}
\label{tab:mul_concept_max}
\centering
\scriptsize
\setlength{\tabcolsep}{4pt}
\begin{tabular}{lccl}
\hline
\textbf{Method} & \textbf{Model} & \textbf{Concept} & \textbf{Top words} \\
\hline
\multirow{6}{*}{ESD} & \multirow{6}{*}{SD1.4} & Church     & wolf, russian, reporter, future, beach \\
                    &                         & Garbage    & blond, blonde, baby, bag, tattoo \\
                    &                         & Parachute  & child, text, conversation, words, daughter \\
                    &                         & Tench      & blond, blonde, russian, artist, professor \\
                    &                         & Van Gogh   & child, text, words, friend, pair \\
                    &                         & Nudity     & target, cross, seal, boyfriend, pattern \\
\hline
\multirow{5}{*}{SPM}          & \multirow{5}{*}{SD1.4} & Church     & church, sermon, house, funeral, priest \\
                    &                         & Garbage    & truck, opera, laundry, bus, musical \\
                    &                         & Parachute  & wind, angel, fairy, comet, balloon \\
                    &                         & Tench      & sun, fish, tree, prince, poet \\
                    &                         & Van Gogh   & painting, painter, poet, picasso, moon \\
                    &                         & Nudity     & body, woman, baby, goddess, sheet \\
\hline
\multirow{5}{*}{SH}           & \multirow{5}{*}{SD1.4} & Church     & map, meeting, room, tray, wall \\
                    &                         & Garbage    & table, room, cast, classroom, meeting \\
                    &                         & Parachute  & band, map, meeting, show, table \\
                    &                         & Tench      & book, calendar, map, meeting, room \\
                    &                         & Van Gogh   & crowd, painting, target, vamp, wall \\
                    &                         & Nudity     & car, camera, horse, play, bear \\
\hline
\multirow{5}{*}{DreamBooth}   & \multirow{5}{*}{SD1.5} & furniture\_sofa  & target, poster, dozen, bible, baby \\
                    &                         & jewelry\_ring     & knife, target, notebook, anthem, laptop \\
                    &                         & plushie\_panda    & text, cat, baby, guy, words \\
                    &                         & things\_book      & target, note, page, text, book \\
                    &                         & toy\_unicorn      & bible, monster, demon, David, target \\
                    &                         & pets\_dog     & baby, tune, owner, body, line \\
\hline
\multirow{5}{*}{DreamBooth}   & \multirow{5}{*}{SD2.1} & furniture\_sofa  & man, face, car, portrait, song \\
                    &                         & jewelry\_ring     & song, flag, towel, book, words \\
                    &                         & plushie\_panda    & picture, dozen, man, voice, guy \\
                    &                         & things\_book      & picture, melody, target, photograph, name \\
                    &                         & toy\_unicorn      & song, picture, target, name, tune \\
                    &                         & pets\_dog     
                       & dog, woman, cat, car, photo \\
\hline
\multirow{5}{*}{DreamBooth}   & \multirow{5}{*}{SD3.5} & decoritems\_lamp  & man, girl, woman, cat, family \\
                    &                         & person          & woman, man, girl, picture, party \\
                    &                         & scene\_lighthouse  & girl, man, face, couple, vampire \\
                    &                         & scene\_waterfall   & girl, couple, vampire, woman, poem \\
                    &                         & transport\_car    & girl, woman, man, child, boy \\
                    &                         & pets\_dog     & bible, dragon, car, report, devil \\
\hline
\end{tabular}
\end{table}

\begin{table}[t]
\caption{Low-drift tokens identified by \method on models adapted to unlearn different concepts. Each row corresponds to an independently adapted checkpoint.}
\label{tab:mul_concept_min}
\centering
\scriptsize
\setlength{\tabcolsep}{4pt}
\begin{tabular}{lccl}
\hline
\textbf{Method} & \textbf{Model} & \textbf{Concept} & \textbf{Top words} \\
\hline
\multirow{6}{*}{ESD}          & \multirow{6}{*}{SD1.4} & Church     & boy, shirt, window, harp, melody \\
                    &                         & Garbage    & house, tree, mansion, pond, mountain \\
                    &                         & Parachute  & city, car, holocaust, church, movie \\
                    &                         & Tench      & house, car, tree, knife, beatles \\
                    &                         & Van Gogh   & house, funeral, piano, choir, mass \\
                    &                         & Nudity     & garden, house, bitch, place, funeral \\
\hline
\multirow{5}{*}{SPM}          & \multirow{5}{*}{SD1.4} & Church     & tune, figure, duck, horse, worker \\
                    &                         & Garbage    & picture, horse, spider, book, photo \\
                    &                         & Parachute  & photo, picture, man, photograph, car \\
                    &                         & Tench      & car, cop, hotel, plane, band \\
                    &                         & Van Gogh   & car, shop, title, bible, package \\
                    &                         & Nudity     & car, war, dog, boat, battle \\
\hline
\multirow{5}{*}{SH}           & \multirow{5}{*}{SD1.4} & Church     & woman, devil, stranger, girl, cross \\
                    &                         & Garbage    & devil, woman, cross, suicide, female \\
                    &                         & Parachute  & female, cross, stranger, soldier, patient \\
                    &                         & Tench      & devil, stranger, cross, woman, female \\
                    &                         & Van Gogh   & man, nun, boy, biography, brother \\
                    &                         & Nudity     & target, shirt, text, bible, pattern \\
\hline
\multirow{5}{*}{DreamBooth}   & \multirow{5}{*}{SD1.5} & furniture\_sofa  & cat, doctor, nurse, baby, train \\
                    &                         & jewelry\_ring     & ship, couple, nurse, man, train \\
                    &                         & plushie\_panda    & restaurant, camera, statue, photo, room \\
                    &                         & things\_book      & house, stranger, shop, woman, life \\
                    &                         & toy\_unicorn      & room, spider, building, night, wall \\
                    &                         & pets\_dog     & landscape, painting, scene, river, girl \\
\hline
\multirow{5}{*}{DreamBooth}   & \multirow{5}{*}{SD2.1} & furniture\_sofa  & tempest, baby, spider, bed, flower \\
                    &                         & jewelry\_ring     & opera, tempest, cat, machine, kitchen \\
                    &                         & plushie\_panda    & scene, novel, bed, stranger, baby \\
                    &                         & things\_book      & painting, party, boy, sermon, mother \\
                    &                         & toy\_unicorn      & bible, scene, tomb, body, bathroom \\
                    &                         & pets\_dog     & face, couple, house, child, comet \\
\hline
\multirow{5}{*}{DreamBooth}   & \multirow{5}{*}{SD3.5} & decoritems\_lamp  & child, woman, girl, vampire, boy \\
                    &                         & person          & girl, woman, car, kid, novel \\
                    &                         & scene\_lighthouse  & woman, girl, man, boy, face \\
                    &                         & scene\_waterfall   & girl, vampire, woman, child, face \\
                    &                         & transport\_car    & girl, man, boy, baby, woman \\
                    &                         & pets\_dog     & couple, man, boy, girl, family \\
\hline
\end{tabular}
\end{table}

\end{document}